\documentclass[runningheads]{llncs}
\usepackage{graphicx}

\usepackage{tikz}
\usepackage{comment}
\usepackage{amsmath,amssymb} 
\usepackage{color}

\usepackage{amsmath}
\usepackage{amssymb}
\usepackage{boldline}

\usepackage{multicol}
\usepackage{booktabs}

\newcommand{\ieours}{\textit{i}.\textit{e}., }

\usepackage[accsupp]{axessibility}  

\usepackage[width=122mm,left=12mm,paperwidth=146mm,height=193mm,top=12mm,paperheight=217mm]{geometry}

\begin{document}
\pagestyle{headings}
\mainmatter

\title{Meta-Sampler: Almost-Universal yet Task-Oriented Sampling for Point Clouds} 

\titlerunning{ } 
\authorrunning{ } 
\author{Ta-Ying Cheng, Qingyong Hu\thanks{Corresponding author}, Qian Xie, Niki Trigoni, Andrew Markham}
\institute{Department of Computer Science, University of Oxford}

\maketitle

\begin{abstract}
Sampling is a key operation in point-cloud task and acts to increase computational efficiency and tractability by discarding redundant points. Universal sampling algorithms (\textit{e.g.,} Farthest Point Sampling)  work without modification across different tasks, models, and datasets, but by their very nature are  agnostic about the downstream task/model. As such, they have no implicit knowledge about which points would be best to keep and which to reject. Recent work has shown how task-specific point cloud sampling (\textit{e.g.,} SampleNet) can be used to outperform traditional sampling approaches by learning which points are more informative. However, these learnable samplers face two inherent issues: \textit{i)} overfitting to a model rather than a task, and \textit{ii)} requiring training of the sampling network from scratch, in addition to the task network, somewhat countering the original objective of down-sampling to increase efficiency. In this work, we propose an \textit{almost-universal} sampler, in our quest for a sampler that can learn to preserve the most useful points for a particular task, yet be inexpensive to adapt to different tasks, models or datasets. We first demonstrate how training over multiple models for the same task (\textit{e.g.,} shape reconstruction) significantly outperforms the vanilla SampleNet in terms of accuracy by not overfitting the sample network to a particular task network. Second, we show how we can train an almost-universal meta-sampler across multiple tasks. This meta-sampler can then be rapidly fine-tuned when applied to different datasets, networks, or even different tasks, thus amortizing the initial cost of training.


\keywords{Point Cloud Sampling, Point Cloud Processing, Meta-Learning}
\end{abstract}

\section{Introduction}

Modern depth sensors such as LiDAR scanners can capture visual scenes with highly dense and accurate points, expanding the real-world applications of point clouds to traditionally challenging 3D vision tasks. However, while existing deep network architectures such as PointNet \cite{pointnet} are capable of consuming these dense point clouds for downstream tasks (\textit{e.g.,} classification, reconstruction), it is standard to downsample initial point cloud to reduce the computational and memory cost, especially for resource-constrained or real-time applications. As such, the objective of extracting a representative subset of points from raw point clouds while maintaining satisfactory performance over various tasks is a key problem.

\begin{figure}[t]
	\begin{center}
		\includegraphics[width=0.95\linewidth]{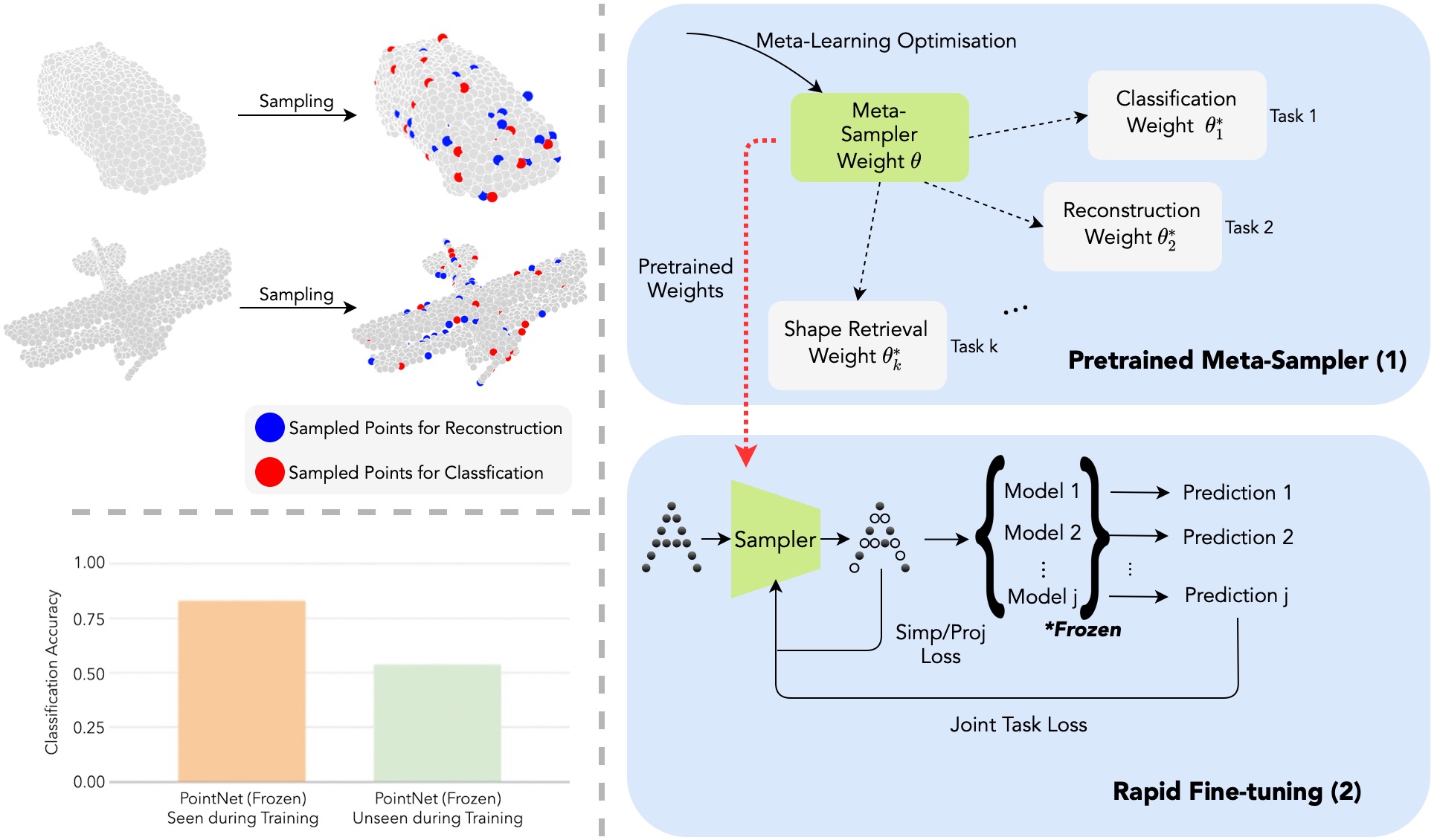}
	\end{center}
	\vspace{-0.3cm}
    \caption{\textbf{Overview.} \textbf{Top Left:} We highlight the points sampled for classification (red) and reconstruction (blue). It is apparent that classification concentrates on more generalised features across the entire point cloud whereas reconstruction focuses on denser aspects for optimisation. \textbf{Bottom Left:} We evaluate the classification performance of two frozen PointNets on 16 sampled points from SampleNet. A large performance gap is observed despite the two models (one adopted during SampleNet training and one unseen) having an identical architecture, implying overfitting onto the model instead of the task itself. \textbf{Right:} An overview of our meta-sampler as a pretrained model that can rapidly adapt with a joint-training mechanism.}
	\label{fig: teaser}
\end{figure}

Early techniques usually adopt Farthest Point Sampling (FPS) \cite{qi2017pointnet++,pointcnn,wu2018pointconv}, Inverse Density Importance Sampling (IDIS) \cite{flex_conv}, or Random Sampling (RS) \cite{hu2019randla,hu2021learning} to progressively reduce the resolution of the raw point clouds. Albeit simple and universal, these sampling schemes are inherently heuristic and task-agnostic. Recently, Dovrat et al. \cite{learning_to_sample} and Lang et al. \cite{samplenet} explored a new domain of learning-based, task-specific, and data-driven point cloud sampling strategies. They empirically proved that leveraging the task loss can effectively optimise the sampler to preserve representative and informative features. Although remarkable progress has been achieved in several downstream tasks such as classification and reconstruction, there remain two critical issues to be further explored: 1) The learnt samplers are shown to \textbf{overfit to a specific task model} instead of being generalisable to the task itself --- this causes a signifant performance drop when adopting another network for the same task even when the two architectures are identical (as exemplified in Figure \ref{fig: teaser} Bottom Left); 2) Training a sampler to fit a particular task is both time-consuming and computationally expensive, which counters the original objective of sampling to improve efficiency.

To this end, we propose an \textit{almost-universal} sampler (Figure \ref{fig: teaser} Right) comprising two training alterations to address the aforementioned issues accordingly. First, we suggest jointly training by forwarding the sampled points to multiple models targetting the same task instead of a single model and updating the sampler through a summation of task losses. This kind of ensemble allows us to better simulate the distribution of different task models, encouraging the sampler to truly learn the task rather than a particular instance. Second, we introduce our meta-sampler to learn how to adapt to a specific task, rather than explicitly learning a particular task model. We incorporate a set of tasks, each with multiple task models, for the meta-optimisation. Our meta-sampler can serve as a pretrained module to adhere to any tasks through fine-tuning while being \textit{almost-universal} in the sense that it could be optimised with fewer iterations.

Extensive experimental results justify the performance and versatility of the proposed meta-sampler. In particular, there is a significant improvement in performance for several mainstream tasks with our joint-training technique compared to the best results from the conventional single-task training on SampleNet. Moreover, we thoroughly evaluate the versatility of our meta-sampler by adapting to particular tasks (both included and excluded from the meta-training), model architectures, and datasets. Our meta-sampler adapts rapidly to all challenging scenarios, making it a suitable pretrained candidate for task-specific learning-based samplers.

In summary, the key contributions of this paper are threefold:
\begin{itemize}
\item A joint-training scheme for the sampler to truly learn a task rather than simply overfitting to a particular instance (\textit{i.e.,} a specific task model).
\item A meta-sampler that can rapidly adapt to downstream point cloud tasks within and beyond the meta-training stage, models of varying architectures, and datasets of different domains.
\item Extensive experiments validate the performance and versatility of our meta-sampler across various tasks, models, and datasets.
\end{itemize}

\section{Related Work}
\subsection{Learning with 3D Point Clouds}
Earlier pursuit of 3D computer vision tasks is mainly focused on grid-like representations of voxel volumes, as the mature convolutional neural networks (CNNs) can be directly extended to such data representation and easily introduce inductive biases such as translational equivariance \cite{3dr2n2,ogn}. However, voxel volume representation has the ingrained drawback of being uniform and low-resolution with densely compacted empty cells consuming vast amount of computational resources. Recently, point-based networks have attracted wide attention with the emergence of PointNet/PointNet++ \cite{pointnet,qi2017pointnet++}. These architectures pioneered the learning of per-point local features, circumvented the constraint of low resolution and uniform voxel representations, and hence introduced significant flexibilities and inspired a plethora of point-based architectures \cite{flex_conv,pointcnn,point_voxel_cnn,dgcnn,point-bert}. A number of point cloud based tasks \cite{guo2020deep} including classification \cite{simpleview,mvtn,point2sequence,pointMLP,walk_in_the_cloud,zhong2022no}, segmentation \cite{semanticKITTI,hu2019randla,superpoint_graphs,hu2022sensaturban,hu2021sqn}, registration \cite{ao2020spinnet,perfect_match,predator}, reconstruction \cite{point_set_gen,3dpsrnet}, and completion \cite{advtrain_upcn,pcn} are extensively investigated. Nevertheless, few arts targeted the fundamental component of point cloud sampling in this deep learning era.

\subsection{Point Cloud Sampling}

Point cloud sampling, a basic component in most point-based neural architectures, is usually used to refine the raw inputs and improve computational efficiency for several downstream tasks. Widely-adopted point cloud sampling methods include RS, FPS \cite{qi2017pointnet++,hough_voting_3D,hu2019randla}, and IDIS \cite{flex_conv}. A handful of recent works began to explore advanced and sophisticated sampling schemes \cite{ahd_sample,fast_resample,grid_gcn}. Nonetheless, despite the remarkable progress in point cloud sampling, these methods are task-agnostic and rather universal, lacking awareness of the important features which a particular task may require.


Recently, Dovrat et al. \cite{learning_to_sample} proposed a learnable, data-driven sampling strategy by imposing a specific task loss to enforce the sampler in learning specific-related features for a particular task. Later, Lang et al. \cite{samplenet} extended the learning approach by introducing a differentiable relaxation to minimise the training and inference accuracy gap for the sampling operation. Nevertheless, by introducing an additional task loss, sampling ultimately becomes constrained and prone to overfitting on a specific task model instead of the task itself. Additionally, this also requires significant extra training to fit a particular goal.

Our meta-sampler hopes to bring the best of both worlds: being task-oriented yet as universal as possible. Instead of directly overfitting onto a task model, we focus on how to learn a task through incorporating a meta-learning algorithm. By introducing a better approach of learning a particular task through joint-training, our pretrained meta-sampler be rapidly fine-tuned to any task, making it \textit{almost-universal} while easing the computational efficiency to which sampling is targeting in the first place.

\subsection{Meta-Learning}
Meta-learning, the process of learning the learning algorithm itself, has shown to be applicable to several challenging computer vision scenarios such as few-shot and semi-supervised classification \cite{maml,meta_semi}, shape reconstruction \cite{few-shot_priors}, and reinforcement learning \cite{meta_rl1,meta-rl2} due to its capacity for fast adaptation.

Finn et al. \cite{maml} proposed one of the most representative meta-learning methods, termed model-agnostic meta-learning (MAML), that allows the model to quickly adapt to new tasks in different domains such as classification, regression, and reinforcement learning. Later, Antoniou et al. \cite{train_maml} further improved the MAML learning scheme, making the learning more generalisable and stable. Recently, Huang et al. \cite{metaSets} proposed MetaSets, which aims to meta-learn the different geometry of point clouds so that the model can generalize to classification tasks performed on different datasets. In contrast and being analogous to the standard meta-learning problem, our proposed meta-sampler focuses on universal point cloud sampling for different tasks, aiming to achieve fast adaptation to reduce computation efficiency through a training strategy extended from \cite{maml,train_maml}. Our fast adaption is not just across tasks within the meta-training, but also across models, datasets and unforeseen tasks.


\section{Meta-Sampler and Rapid Task Adaptation}

Ideally, it is desirable to learn a unified and universal point cloud sampling scheme for different tasks in a data-driven manner --- this is most likely unfeasible since different tasks inherently have distinctive preferences of sampling strategies, as shown in Figure \ref{fig: teaser} top left (for more qualitative comparisons please refer to the Appendices). For example, 3D semantic segmentation pays more attention to the overall geometrical structure, while 3D object detection naturally puts more emphasis on the foreground instance with sparse points \cite{zhang2022not}. Motivated by this, we take the next-best objective, which is to learn a highly adaptive sampling network that can adapt to a number of tasks with minimal iterations and achieve optimal performances. In particular, this fast adaptation capability allows samplers to be pretrained then quickly fine-tuned, satisfying the ultimate goal of improving computational efficiency.

\subsection{Problem Setting}
\label{Sec3.1}
The goal of this paper is to develop a learning-based sampling module $f_\theta$ with trainable parameters $\theta$, which takes in a point cloud with $m$ points and outputs a smaller subset of $n$ points ($m > n$). Apart from the objective of SampleNet to learn task-specific sampling (\textit{i.e.,} particularly suitable for a single task such as shape classification or reconstruction), we take a step further and aim to propose an universal pretrained model, which can be rapidly adapted to a set of different tasks $S_T = \{T_i\}_{i=1}^{K_T}$.  Formally, we define the ideal adaptation of sampling to a specific task $T_i$ as capable of achieving satisfactory performance by integrating the sampling module into a set of known networks $S_{A_i} = \{A_{i,j}\}_{j=1}^{K_{A_i}}$ (trained with unsampled point clouds of $m$ points to solve task $T_i$. ). We split $S_{A_i}$ into $S_{A_i}^{train}$ and $S_{A_i}^{test}$ (\textit{i.e.,} task networks used during training are disjoint to the ones for testing) to make sure that our evaluation on $f_\theta$ is fair and not overfitting to task models instead of the task itself. Note that while $S_{A_i}^{train}$ is available during training, the weights are frozen when learning our sampler as suggested by \cite{samplenet}.


To achieve the dual objectives of high accuracy and rapid convergence, we must first carefully evaluate the best training strategy to better learn each individual task, and then design a training strategy which is adaptive to multiple tasks. We build our sampler $f_\theta$ based on the previous state-of-the-art learnable sampling network --- PointNet-based SampleNet architecture \cite{pointnet,samplenet} --- and then introduce our training technique in a bottom-up manner.

\subsection{Single-Task Multi-Model Training} \label{joint_task}
For an individual task $T_i$, we hope that the $f_\theta$ learns to sample the best set of points $\forall A \in S_{A_i}$.

The conventional way of training the SampleNet uses one single frozen network $A'$ as $S_A^{train}$ for training by defining a sampling task loss $\mathcal{L}_{ST_i}$ targeting $T_i$ as:

\begin{equation}
    \mathcal{L}_{ST_i}(f_\theta) = \mathcal{L}_{T_i}(A'(f_\theta)),
    \label{eqn:loss_single_task}
\end{equation}
where $\mathcal{L}_{T_i}$ is the loss when pretraining $A'$. We refer to this configuration as single-model, single-task training.
As mentioned previously, this training method, having accomplished promising results in several tasks, still exhibits a large accuracy discrepancy between the results on $A'$ and $S_A^{test}$. In other words, even though $A'$ is frozen during the training of SampleNet, the sampling stage is overfitted onto the task network instead of the task itself.

To alleviate the issue of model-wise overfitting, we extend (\ref{eqn:loss_single_task}) and create a joint-training approach for a single task. Specifically, we take a set of weight-frozen models $\{A_{i,j}\}_{j=1}^{k}, 1 < k << K_{A_i}$ as $S_{A_i}^{train}$ and compute $\mathcal{L}_{ST_i}$ as:

\begin{equation}
    \mathcal{L}_{ST_i}(f_\theta) = \sum_{j=1}^{k}\mathcal{L}_{T_i}(A_{i,j}(f_\theta)).
    \label{eqn:loss_multi_task}
\end{equation}

It is critical to understand that all the frozen task models are under inference mode (\textit{i.e.,} not significantly sabotaging computation power) and that only a very small number of task models (easily obtainable online or by  self-training with different random initial weights) would bring significant improvements to the sampler's performance. We further show in Section \ref{sec:single_task} that a very small $k > 1$ allows the sampling network generalise better across $S_{A_i}$, as $S_{A_i}^{train}$ becomes a vicinity distribution rather than a single specific instance to $S_{A_i}$. 

In addition to the joint $\mathcal{L}_{ST_i}$, we also update the weights with a simplification loss comprising the average and maximum nearest neighbour distance and a projection loss to enforce the probability of projection over the points to be the Kronecker delta function located at the nearest neighbour point (identical to the SampleNet loss \cite{samplenet}).

\subsection{Multi-Task Multi-Model Meta-Sampler Training} \label{Sec3.3}

Instead of restricting ourselves to a single task (\textit{e.g.,} classification), we consider whether training the sampler over multiple tasks could lead to our vision of an almost-universal sampler. Broadly, we aim to extend the sampler beyond multi-model to multi-task, such that given any task $T_i \in S_T$, where $S_T$ is a set of tasks, a good initial starting point could be achieved for the sampler. In this way, adapting or fine-tuning to a particular task (which may even be beyond the known set) will be rapid and cheap.


To tackle this, we draw inspiration from the MAML framework and propose a meta-learning approach for rapidly adaptive sampling \cite{maml}. In essence, we aim to utilise the set of $S_{A_i}^{train}$ to mimic the best gradients in learning a particular task $T_i$ for meta-optimisation, such that given any task $T_i \in S_T$ or even beyond the known set of tasks, the MAML network can quickly converge within a few iterations and without additional training of the task networks.

The joint-training procedure discussed in the previous section motivates that a particular task is better solved with a set of task networks instead of just one --- we transfer this idea to the meta-optimisation such that the sampler is adaptive to a number of tasks instead of just one. Formally, we first optimise the adaptation of $f_\theta$ to $T_i \in S_T$ by updating the parameters $
\theta$ to $\theta'_{i,j}$ for every $A_{i,j}$ through the gradient update:
\begin{equation}
    \theta'_{i,j} = \theta - \alpha\nabla\mathcal{L}_{T_i}(A_{i,j}(f_\theta)),
    \label{eqn:meta_inner_update}
\end{equation}
 where $\alpha$ is the step size hyperparameter. Similar to MAML, we can directly extend the single gradient update into multi-gradient updates to optimise the effectiveness of $\theta'_{i,j}$ on $T_i$.
 
With the inner update (\ref{eqn:meta_inner_update}), we then follow the meta-optimisation procedure through a stochastic gradient descent:
\begin{equation}
    \theta = \theta - \beta\nabla\mathcal\sum_{i=1}^{K_T}\sum_{j=1}^{k}\mathcal{L}_{T_i}(A_{i,j}(f_{\theta'_{i,j}})),
    \label{eqn:meta_update}
\end{equation}
where $\beta$ is the meta step size hyperparameter that could either be fixed or accompanied with annealings. Note that we apply the single task loss in the inner update (\ref{eqn:meta_inner_update}) but sum all losses from all weights to resemble a task in the meta-update (\ref{eqn:meta_update}). Section \ref{versatility} shows that our meta-optimisation design is sufficient in learning tasks for rapid adaptation. Simplification and projection losses are also directly optimised at this stage. They are however directly updated rather than included in the meta-update fashion as they are task-agnostic.

\begin{figure}[htb]
	\begin{center}
		\includegraphics[width=1.0\linewidth]{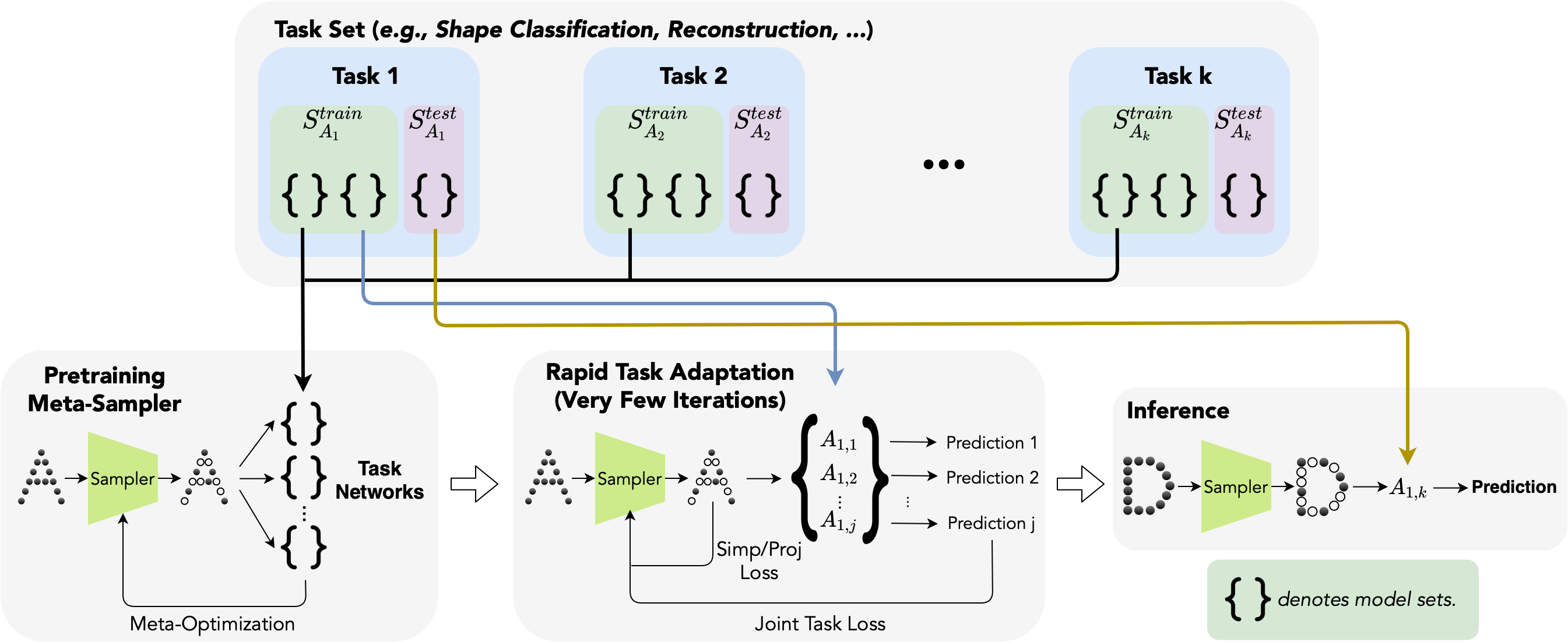}
	\end{center}
    \caption{\textbf{The pipeline of the proposed meta-sampling.} The illustration exemplifies the pretraining with multiple tasks through our meta-training strategy, then fitting onto a single task with our joint-training mechanism.}
	\label{fig: pipeline}
\end{figure}
\subsection{Overall Pipeline: Pretrained Meta-Sampler to Task Adaptation} \label{sec3.4}
We describe the overall training pipeline of the proposed meta-sampler (Figure \ref{fig: pipeline}) as the following:

\smallskip\noindent\textbf{Pretrained Meta-Sampler: } Our pipeline begins with training a meta-sampler. First, we take a set of tasks $S_T$ (\textit{e.g.,} shape classification, reconstruction, retrieval) and their corresponding task networks $S_{A_i}$ for every $T_i\in S_T$ (pretrained on the unsampled point clouds). Next, we freeze all their original weights and perform our meta-sampler training as illustrated in Section \ref{Sec3.3} to obtain a pretrained meta-sampler.

\smallskip\noindent\textbf{Rapid Task Adaptation: }The meta-training attempts to optimise $\theta$ to a position optimal to learn any task $T_i$. Therefore, to adapt to a particular task, we can simply take the pretrained weights of the meta-sampler and fine-tune it with the joint-training strategy as illustrated in \ref{joint_task} along with the previously proposed simplification and projection loss.

\smallskip\noindent\textbf{Disjoint Task Networks for Pretraining and Training: } Realistically, one should be able to directly obtain a pretrained meta-sampler without the task networks and fit to their own networks. To mimic such real-world constraints, we ensure that the meta pretraining and joint-training use disjoint sets of networks --- both of which are unseen during testing.

\section{Experiments}

Our empirical studies comprise two major components. First, we evaluate the performance of the proposed joint-training scheme against prior training methodologies on representative individual tasks. Afterward, we justify the versatility and robustness of the meta-sampler by measuring its adaptiveness across different tasks, models, and datasets.

\subsection{Experimental Setup}

To comprehensively evaluate the performance of our meta-sampler, we extract a set of representative tasks on 3D point clouds, including shape classification, reconstruction, and shape retrieval. Note that all experiments are conducted on the ModelNet40 \cite{modelnet40} (except for ShapeNet  \cite{chang2015shapenet} used in transferring dataset analysis) to ensure fair evaluation (\textit{i.e.,} without introducing additional information during meta-sampler training). The detailed experimental settings (\textit{i.e.,} task network architecture, task loss) are described as follows:

\smallskip\noindent\textbf{Shape classification.} This is a fundamental task in 3D vision to determine the shape categories of a given point cloud. The task network set $\{A_{i,j}\}$ are pretrained PointNets \cite{pointnet} with random and distinct weight initialisations, and the validation accuracy converges to 89\% to 90\%. $\mathcal{L}_{T_i}$ is the vanilla binary cross-entropy (BCE) loss for classification.

\smallskip\noindent{\textbf{Reconstruction.} This task aims to reconstruct the complete 3D shape from partial point sets. Following \cite{samplenet}, the goal of sampling for this task is to preserve $n$ key points that could be reconstructed to the original unsampled point clouds. For this task, $S_{A_i}$ is a set of Point Completion Networks (PCN) \cite{pcn} trained in an autoencoder fashion to minimise the Chamfer Distance (CD) between the input and output points. We select the PCN architecture owing to its encoder and decoder mechanism that doesn't take in any structural assumptions (\textit{e.g.,} symmetry), making it suitable for reconstruction even when missing points are randomly distributed upon the entire shape instead of a particular part. $\mathcal{L}_{T_i}$ is the two-way CD between the inputs and predicted outputs; all networks are pretrained with the loss to the chamfer distance of around $3\times 10^{-4}$.}

\smallskip\noindent{\textbf{Shape Retrieval.} Given a sampled point cloud, the goal is to match it with the shifted/rotated original point cloud given $N$ options (similar to the $N$-way evaluation in few-shot settings). Due to the existence of hard negative pairs (point clouds of the same class), this task requires more advanced learning of fine-grained features compared with the pure shape classification. In this case, $S_{A_i}$ is a set of Siamese PointNets inspired by \cite{koch2015siamese} pretrained on unsampled point clouds matching. $\mathcal{L}_{T_i}$ is a BCE loss where the ground truth is set to 1 if the point cloud is a shifted/rotated version of the other and vice versa. All networks are pretrained to achieve $100\%$ accuracy on the simple 4-way evaluation. }

\subsection{Performance Evaluation on Individual Tasks} \label{sec:single_task}

\begin{table}[t]
  \centering
  \resizebox{0.99\columnwidth}{!}{%
    \setlength\tabcolsep{2pt} 
    \begin{tabular}{ccccc}
    \toprule
    & \multicolumn{4}{c}{\textbf{Classification (Accuracy $\uparrow$)}} \\
    \cmidrule(lr){2-5}
     \textbf{Sampling Ratio}  & FPS & SNet\cite{learning_to_sample} & Single\cite{samplenet} & Joint\\
    \cmidrule(lr){2-5}
    8   &   70.4\% & 77.5\%  & 83.7\%     &  \textbf{88.0\%}        \\
    16  &   46.3\% & 70.4\%  & 82.2\%     &  \textbf{85.5\%}         \\
    32  &   26.3\% & 60.6\%  & 80.1\%     &  \textbf{81.5\%}      \\
    64  &   13.5\% & 36.1\%  & 54.1\%     &  \textbf{61.6\%}        \\
    \bottomrule
    \\
    \end{tabular}
    \quad
    \begin{tabular}{ccc}
    \toprule
    & \multicolumn{2}{c}{\textbf{Reconstruction (CD $\downarrow$)}} \\
    \cmidrule(lr){2-3} 
     \textbf{Sampling Ratio}  & Single\cite{samplenet} & Joint  \\
    \cmidrule(lr){2-3} 
    8   &   3.29      &     \textbf{3.05}    \\
    16  &   3.32      &     \textbf{3.15}     \\
    32  &   3.61      &     \textbf{3.37}    \\
    64  &   4.43      &     \textbf{4.31}     \\
    \bottomrule
    \\
    \end{tabular}
    }
    \resizebox{0.75\columnwidth}{!}{%
    \setlength\tabcolsep{2pt} 
    \begin{tabular}{ccccccc}
    \toprule
    & \multicolumn{6}{c}{\textbf{Shape Retrieval (Accuracy $\uparrow$)}} \\
    \cmidrule(lr){2-7}
    & \multicolumn{2}{c}{\textbf{4-way}} & \multicolumn{2}{c}{\textbf{10-way}}
   & \multicolumn{2}{c}{\textbf{20-way}}\\
    \cmidrule(lr){2-3}\cmidrule(lr){4-5}\cmidrule(lr){6-7}
     \textbf{Sampling Ratio} & Single\cite{samplenet} & Joint & Single\cite{samplenet} & Joint & Single\cite{samplenet} & Joint\\
    \cmidrule(lr){2-3}\cmidrule(lr){4-5}\cmidrule(lr){6-7}
    8   &   99.6\% & \textbf{99.7\%}  & 96.3\% & \textbf{98.3\%} & 95.9\% & \textbf{96.7\%}     \\
    16  &   98.7\% & \textbf{99.1\%}  & 94.0\% & \textbf{96.7\%} & 89.5\% & \textbf{91.9\%}      \\
    32  &   97.2\% & \textbf{97.5\%}  & 91.4\% & \textbf{91.5\%} & 82.9\% & \textbf{84.6\%}    \\
    64  &   92.5\% & \textbf{94.6\%}  & 79.5\% & \textbf{84.6\%} & 67.0\% & \textbf{71.0\%}     \\
    \bottomrule
    \\
    \end{tabular}
    }
    \\
    \caption{\textbf{Joint v.s. Single Task Network Training.} Single and Joint denotes the SampleNet trained through the originally proposed single task network approach \cite{samplenet} and through our proposed multi-model single-task training ($k=3$), respectively. Classification and Shape Retrieval performances are measured directly with accuracy (higher is better). Reconstruction performance is measured with Chamfer Distance at a scale of $10^{-3}$ (lower is better). Bold texts denote best results.}
    \label{Table1}
    
    \end{table}


To justify the effectiveness of the proposed multiple-model training scheme, we present the quantitative comparison of incorporating multiple models training and the traditional SampleNet single-model training strategy in all three individual tasks on the ModelNet40 dataset \cite{modelnet40}. We adopt the official train and test splits in this dataset, and follow \cite{learning_to_sample,samplenet} to pretrain all task networks with the original point clouds (1024 points by default). All the task models are under inference mode during the training of the sampling network. We evaluate our sampling on different sampling ratios calculated as $1024/n$, where $n$ is the number of outputted points from the sampler. Note that for reconstruction and shape retrieval tasks, we were adopting a different dataset and task to prior networks. Much work is required for the adaption and thus we only compare with the previously proposed state-of-the-art SampleNet.

\smallskip\noindent\textbf{Shape Classification.} As shown in Table \ref{Table1}, the classification performance achieved with our joint-training scheme consistently outperforms the single SampleNet and previous sampling strategies such as FPS across all sampling ratios. In particular, the classification accuracy achieved with our joint-training mechanism under a sampling ratio of 8 is very close to the upper bound accuracy (88.0\% vs. 89.5\%) achieved without any sampling, verifying the effectiveness of our joint-training strategy. We also notice that the performance gap between the proposed method and others is widening with a growingly aggressive sampling rate (\textit{e.g.,} at sampling ratio 64 with 16 points left for the point clouds). This further demonstrates the superiority of the proposed training mechanism in preserving task-significant features.

\smallskip\noindent\textbf{Reconstruction.} The effect of joint-training on reconstruction follows a similar trend to shape classification, outperforming other strategies in terms of the CD across all sampling ratios. The improvement seems to be consistent across all sampling ratios, further exhibiting the effectiveness of joint-training.

\smallskip\noindent\textbf{Shape Retrieval.} Our shape retrieval results are presented under the \textit{N}-way few-shot settings (\textit{N}$=4, 10, 20$). It is clear that the joint-training scheme achieves better results compared with the single SampleNet training strategy. Specifically, the advantages of our proposed joint-training scheme is more prominent with the increase of the sampling ratios, suggesting that our sampling schemes can preserve points that have high similarity to the original point cloud.


\smallskip\noindent\textbf{Faster Convergence with multiple models.} Considering the sampler is exposed to more task networks during training, it is expected for the sampler to converge to stabilised accuracies with a shorter time span of training. Our empirical study generally aligns with this idea. Our joint-training taking usually around 40 epochs to converge as opposed to around 60 for single-task training.

\begin{table}[t!]
\centering
\begin{tabular}{p{2cm}m{1cm}m{1cm}m{1cm}m{1cm}m{1cm}}
\toprule
& \multicolumn{5}{c}{\textbf{Number of Task Models ($k$)}} \\
\cmidrule(lr){2-6} 
&\textbf{$k=1$}&\textbf{$k=2$}&\textbf{$k=3$}&\textbf{$k=4$}&\textbf{$k=5$}\\
\cmidrule(lr){2-6}
\textbf{Accuracy} & 80.0\% & 81.3\% & 81.7\% & 82.8\% & 83.4\%\\
\bottomrule
\\
\end{tabular}
\caption{\textbf{Classification Accuracy when Increasing $k$}. Sampling Ratio is 32. }
\label{tab:increase_k}
\end{table}

\smallskip\noindent\textbf{The Impact of the Number of Task Models $k$.} We further dive into the correlation between the number of networks $k$ for joint-training and classification accuracy. Table \ref{tab:increase_k} shows the classification results under the randomly selected sampling ratio of 32 when we slowly increase the number of task networks for ensemble. A clear trend of increments continues as $k$ increases, implying that the wider the set of training networks the better the approximation is to the entire task distribution. Nonetheless, such increments suffer from the trade-off in computational resources time and memory-wise. We stick with 3 networks as the standard for joint-training and in later experiments unless specified.


\vspace{-0.5cm}

\subsection{Versatility of Meta-Sampler} \label{versatility}
Versatility is a broad term with multiple dimensions requiring evaluation. To fully realize this, we begin with the critical evaluation of our meta-sampler's adaptiveness on tasks included in the meta-optimisation step. We then extend to the more challenging scenarios of changing model architecture, dataset distribution, and ultimately tasks distinct from the ones used for meta-training. All experiments are conducted with hyperparameters $\alpha$ and $\beta$ set to 1e-3.

\begin{figure}[htb]
	\begin{center}
		\includegraphics[width=1.0\linewidth]{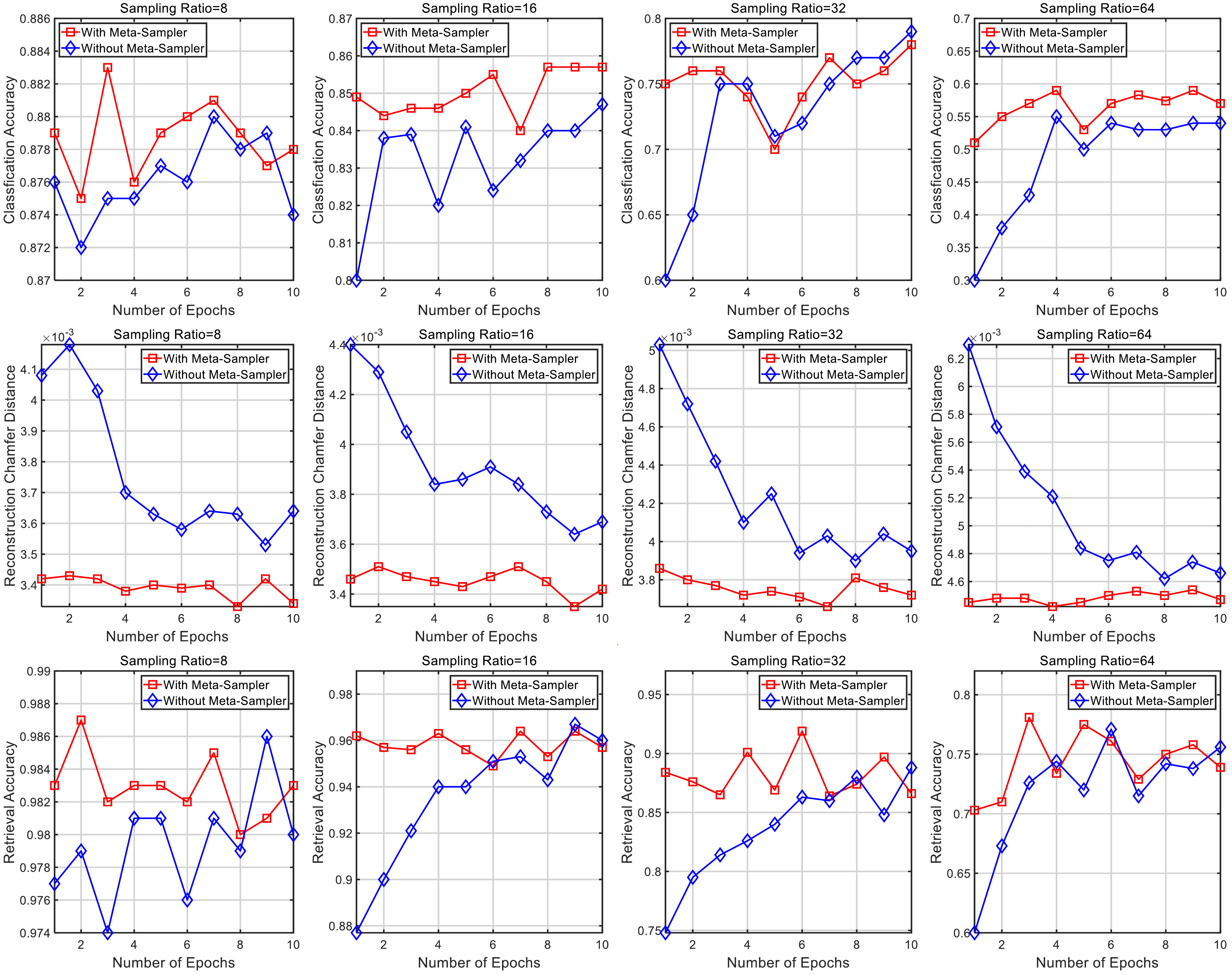}
	\end{center}
    \caption{\textbf{The performance comparison in classification (row 1 $\uparrow$), reconstruction (row 2 $\downarrow$), and shape retrieval (row 3 $\uparrow$) with/without the meta-sampler at the initiation of training.} All graphs begin after one epoch. Red is ours.}
	\label{fig: combine}
\end{figure}

\smallskip\noindent\textbf{Converging to meta-tasks.}
To investigate the impact of the meta-sampler for the performance of meta tasks, we conduct several groups of experiments in this section for the three tasks used in our meta-optimisation, including shape classification, reconstruction, and shape retrieval. We compare the task performance achieved with/without the pretrained meta-sampler under different sampling ratios in Figure \ref{fig: combine}. Specifically, for our meta-sampling, we first deliberately select a bunch of task models unseen during meta-optimisation to fine-tuned the meta-sampler with the joint-training scheme, then evaluate the task performance with the sampled point clouds.

As shown in Figure \ref{fig: combine}, we separately compare the task performance as the training progresses with/without our pretrained meta-sampler for different meta-tasks.  It is clear that as the sampling ratio increases (\ieours the task is more challenging), joint-training without meta-sampler starts at lower performance and requires more iterations to converge to a stable result. By contrast, our pretrained meta-sampler allows the network to quickly adapt to the task within one epoch across all sampling ratios and achieves higher accuracies after 10 epochs in most cases.



There are also two intriguing points we would like to address within this empirical study. First, we observe a relatively large fluctuation of the performance for shape classification and shape retrieval --- a phenomenon we conjecture to be owing to the distribution shift between training and testing sets. Second, we notice that our meta-sampler not only converges faster, but also pushes the upper bound in some cases. For example, our shape retrieval results at sampling ratio 16 achieved the accuracy of 96.9\% (the upper bound of training from scratch is 96.7\%) on the 20th epoch (not plotted in the figure). This infers that by learning how to adapt, the sampling model could potentially also be trained to learn better. However, such occurrences do not take place at all times.

\begin{figure}[t]
    \centering
  \includegraphics[width=0.6\linewidth]{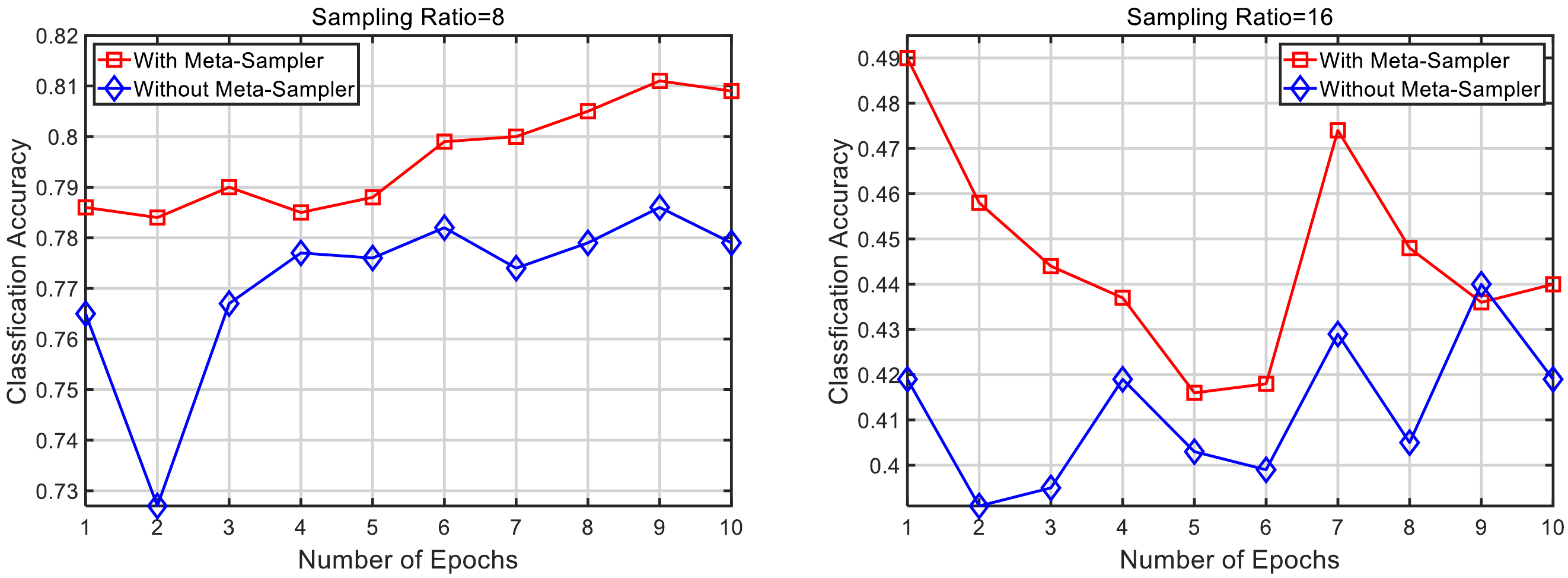}
  \caption{\textbf{Accuracy of PointNet++ for Classification with joint-training}. The results with and without the pretrained meta-sampler on PointNet is presented for sampling ratios at 8 and 16. All graphs begin after one epoch. }
  \label{fig:transfer_models}
\end{figure}

\smallskip\noindent\textbf{Transferring beyond model architectures.} Prior experiments focus on training from the meta-sampler with task networks of identical architecture but different weights. To further explore the versatility of our meta-sampler, we transfer the joint-training networks from PointNets to PointNet++ \cite{pointnet,qi2017pointnet++}. All the networks are pretrained until convergence (\ieours around 92\% accuracy on unsampled point clouds). Constrained by the original implementation of PointNet++ (\ieours point set abstraction layer) extracting features from the 32 points neighborhoods, we only evaluate our meta-sampler upon the sampling ratio of 8 and 16, where the remaining point cloud size is greater than 32.

\begin{figure}[t]
    \centering
  \includegraphics[width=\linewidth]{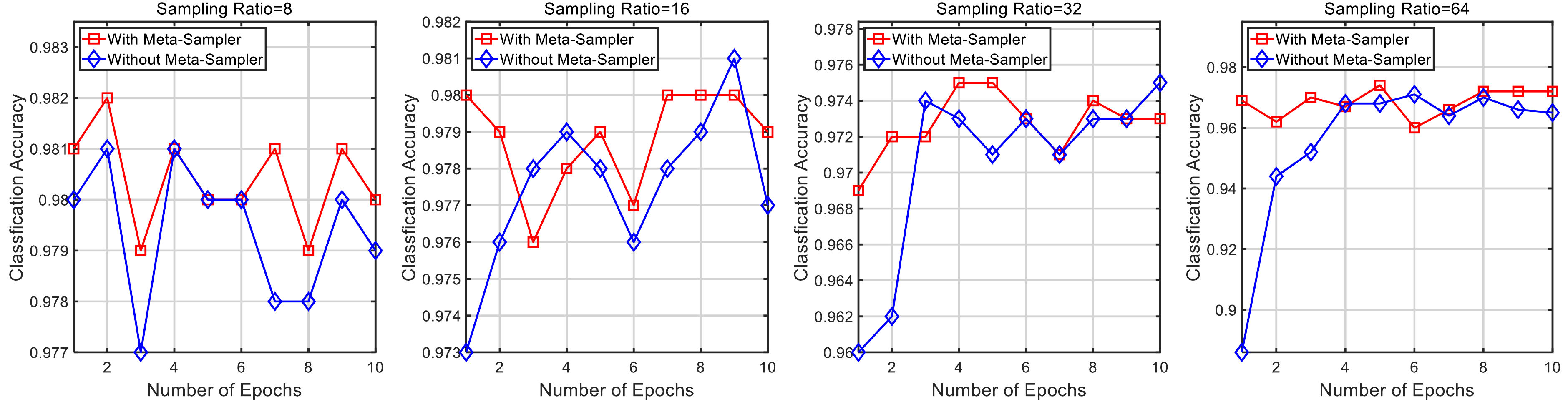}
  \caption{\textbf{Transfer to ShapeNet}. We adopt the ModelNet pretrained meta-sampler to fit onto the ShapeNet dataset for classification. All graphs begin after one epoch. }
  \label{fig:transfer}
\end{figure}

The achieved results are plotted in Figure \ref{fig:transfer_models}. It is apparent that better performance is achieved using our meta-sampler, with a higher starting point and fast convergence speed under both sampling ratios. This further demonstrates the capacity of our meta-sampler in adapting different model architectures. Interestingly, we also notice that the classification accuracy of PointNet++ achieved on sampled points drops significantly compared with that of unsampled point clouds, especially under the sampling ratio of 16. This is likely because FPS is progressively used in each encoding layer. In this case,  by adding a SampleNet in front of PointNet++, we are implicitly ``double sampling" and leaving very few features for the abstraction layer.

\smallskip\noindent\textbf{Transferring Datasets.} To verify that our ModelNet40 \cite{modelnet40} pretrained sampler isn't applicable to just the data distribution it was expose to, we measure the effectiveness when using the same pretrained model for the same task but on a different dataset. Specifically, we evaluate the classification performance of our meta-sampler on the ShapeNetCoreV1 Dataset \cite{chang2015shapenet}, which comprises point cloud objects from 16 different shape categories. Specifically, we still adopt the PointNet \cite{pointnet} architecture and trained three networks following the best practise, while these models can achieve around 98\% accuracy on unsampled point clouds. We then show the training progress achieved by using our joint training scheme with/without the meta-sampler. As shown in Figure \ref{fig:transfer}, although the performance is similar when the sampling ratio is small (easier), we can clearly notice that the model without our meta-sampler starts at a much lower performance. By contrast, the model with our pretrained meta-sampler converges much faster (even within one epoch) and is more stable. As such, this empirical study can well prove that our meta-sampler can adapt to a completely disparate dataset distribution and serve as a better and more stabilised starting point.

\smallskip\noindent\textbf{Beyond Meta Training Tasks.} Finally, we extend to the most challenging question of whether the proposed meta sampler can generalize to unseen tasks, i.e. tasks that are not included in the meta-optimisation step. This is highly challenging since different tasks inevitably have distinct preferences in sampled points. However,  this is also a critical step to validate whether the proposed meta sampler could be the universal pretrained module for all point cloud tasks.

\begin{figure}[t]
	\begin{center}
		\includegraphics[width=0.6\linewidth]{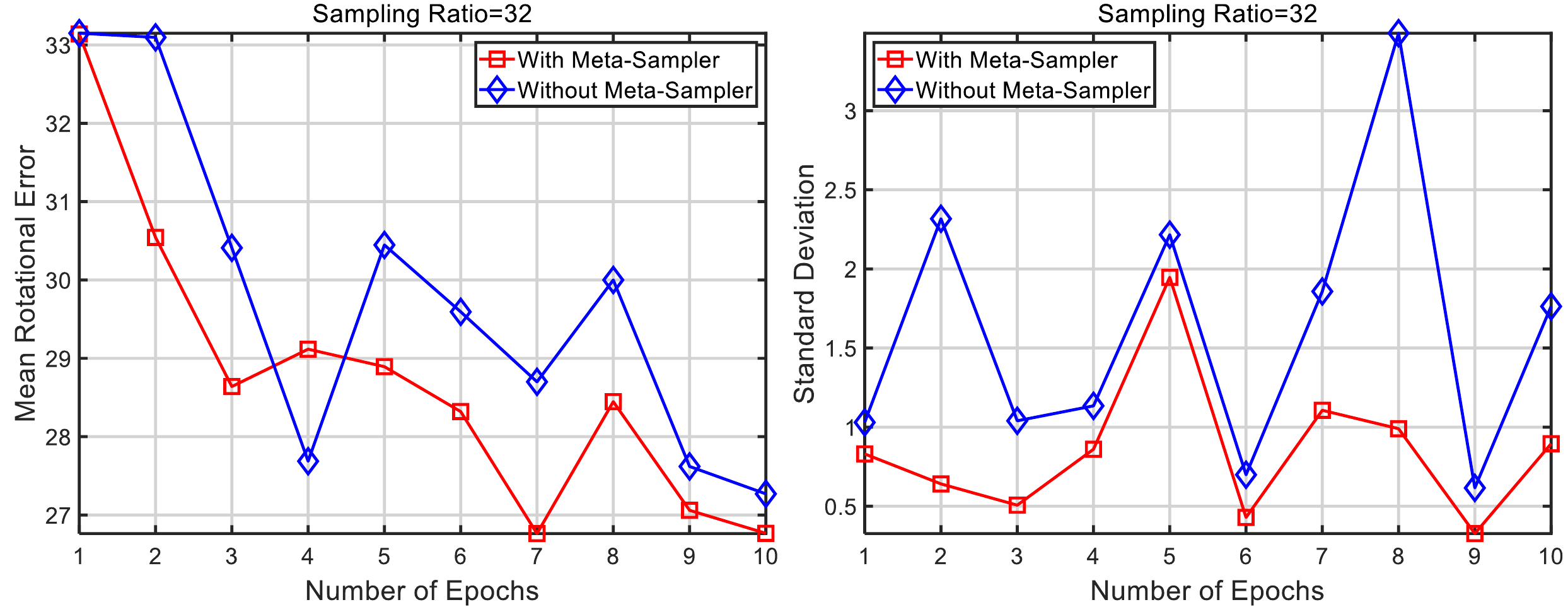}
	\end{center}
    \caption{\textbf{The performance comparison in point cloud registration with/without the meta-sampler.} \textbf{Left:} Rotational error comparison. Right: Standard deviation of rational error per epoch. All graphs begin after one epoch. }
	\label{fig: ex_registration}
\end{figure}


We evaluate our meta-sampler on the point cloud registration --- the task of finding the spatial transformation between two point clouds. Here, we follow the standard train-test split of PCRNet \cite{vsarode2019pcrnet} to obtain pairs of source and template point clouds with templates rotated by three random Euler angles of $[-45^{\circ}, 45^{\circ}]$ and translated with a value in the range $[-1, 1]$. $\mathcal{L}_{T_i}$ is the CD between the source point cloud and the template point cloud with our predicted transformation. We first train three PCRNets to achieve the rotation error of around 7-9 degrees on unsampled point clouds, then freeze the PCRNet weights and perform the proposed joint-training scheme under the conditions with/without the pretrained meta-sampler under the sampling ratio of 32. We ran each setting three times and show the mean and standard deviation of rotational error during training in Figure \ref{fig: ex_registration}. Even though the task objective (registration), task network (PCRNet), and even the dataset itself (pairs of point clouds from ModelNet40 with extra transformations) are unforeseen during our meta-optimisation, we can still notice two subtle yet solid performance differences adopting our meta-sampler: 1) The pretrained model generally converges faster during the initiation of training and 2) The pretrained model is much more stabilised and improves consistently compared to the model trained from scratch that exhibits a large variance throughout every epoch.

\section{Conclusion}
We propose a learnable meta-sampler and a joint-training strategy for task-oriented, almost-universal point cloud sampling. The proposed multi-model joint training scheme on SampleNet achieved promising performance for various point cloud tasks, and the meta-sampler has empirically shown to be effective and stabilising when transferred to any tasks incorporated during meta-optimisation, even extending to unseen model architectures, datasets, and tasks. We hope our pretrained meta-sampler can be used as a plug-and-play module and widely deployed to point cloud downstream tasks to save computational resources.

\clearpage
%
%
\bibliographystyle{splncs04}
\bibliography{egbib}
\end{document}


\pagestyle{headings}
\mainmatter
\def\Number{1516}  

\title{Appendices}

\titlerunning{Meta-Sampler} 
\authorrunning{Meta-Sampler} 
\author{Ta-Ying Cheng, Qingyong Hu, Qian Xie, Niki Trigoni, Andrew Markham}
\institute{Department of Computer Science, University of Oxford}

\maketitle

\section{Limitations}
While this work shows the effectiveness of incorporating multiple tasks in the meta-training step, the question of finding the best combination and the number of tasks for the best model is indeed intriguing and important. Intuitively, more tasks will supposedly lead to more exposure to different task varieties. Simultaneously, this could also lead to tasks potentially contradicting each other and thus diminishing the effectiveness of the pretrained meta-sampler.

\begin{figure}[ht]
	\begin{center}
		\includegraphics[width=0.95\linewidth]{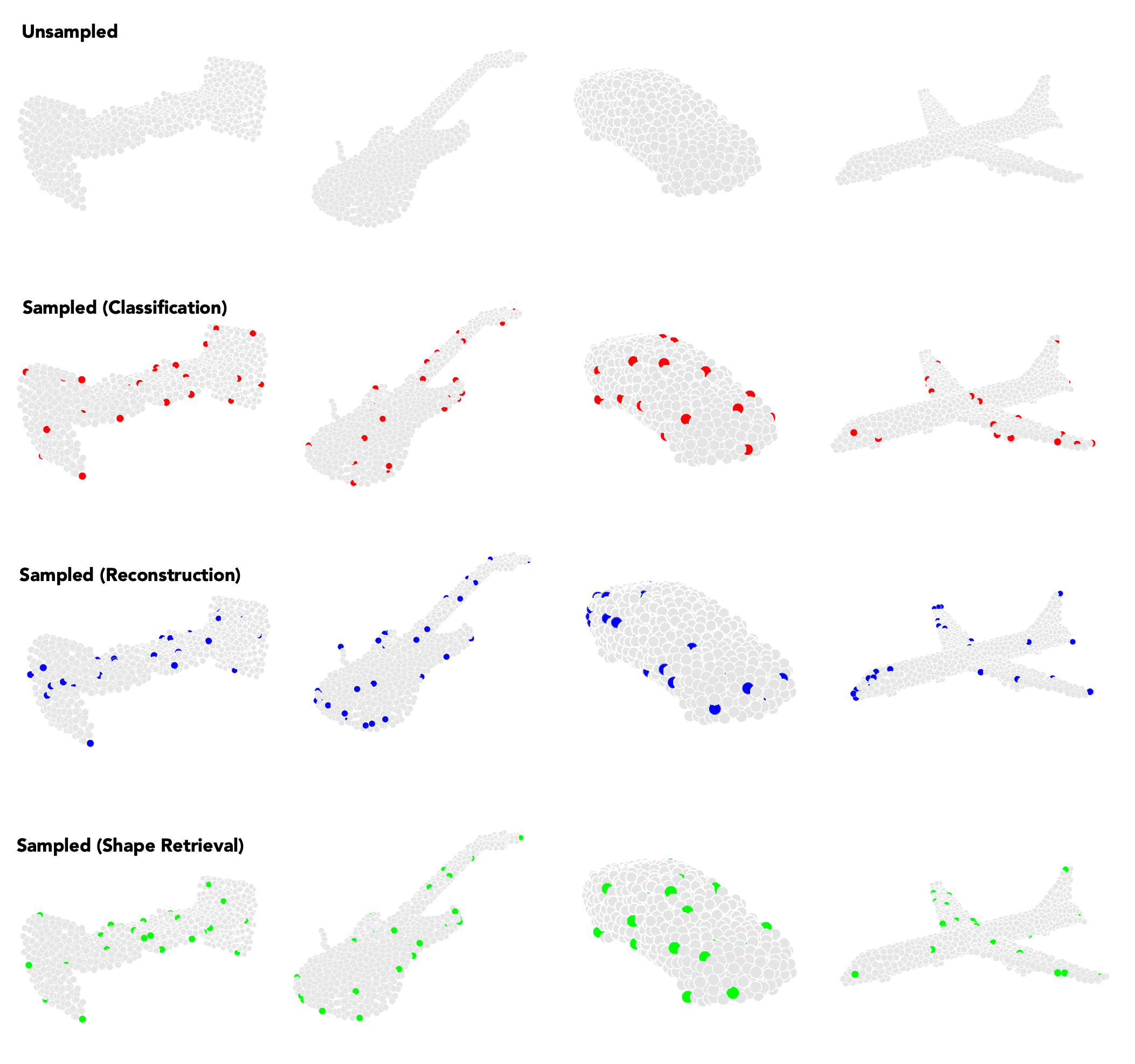}
	\end{center}
	\vspace{-0.1cm}
    \caption{\textbf{Visualisations of sampled points for classification (red), reconstruction (blue), and shape retrieval (green) at sampling ratio of 32.}}
	\label{fig: vis}
\end{figure}

\begin{figure}[t]
	\begin{center}
		\includegraphics[width=0.8\linewidth]{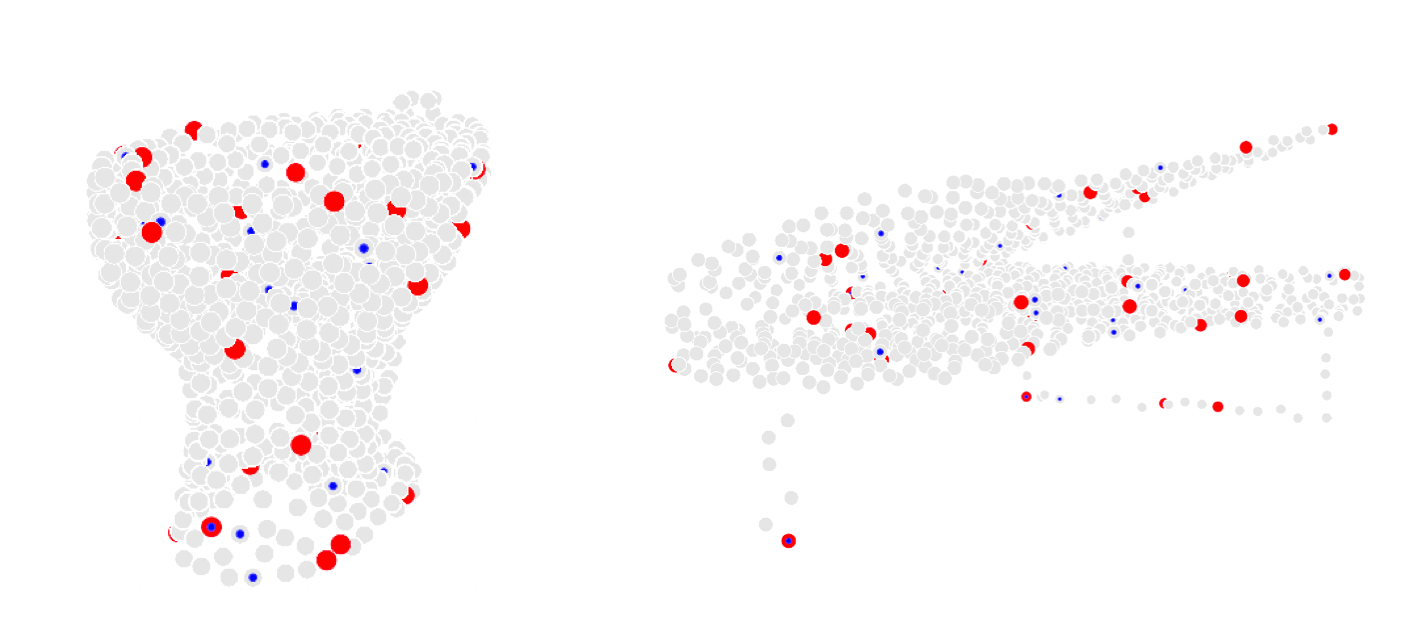}
	\end{center}
	\vspace{-0.1cm}
    \caption{\textbf{Visualisations for SampleNet trained with single-task single-model (blue) and with single task multi-model (red) at sampling ratio of 32.}}
	\label{fig: vis2}
\end{figure}


Another interesting direction is the sampling strategy for tasks such as large-scale segmentation, where the conventional sampling approaches are progressively used in the network, instead of simply using it once at the beginning. How to adapt our meta-sampler to such tasks efficiently and effectively is worth exploring.

\section{Qualitative Results}

\subsection{Sampling for Different Tasks}
We provide additional qualitative analyses on the different points sampled from different tasks for classification, reconstruction, and shape retrieval (Figure \ref{fig: vis}). The results further validate our assumption that different tasks have different preferences in terms of feature sampling (\textit{e.g.,} classification often require the sampled points to be scattered to give an overview of the object; reconstruction focuses on denser points to minimise the Chamfer Distance loss during reconstruction; shape retrieval often preserve fine-grained features like corners to distinguish hard negative of the same class), and hence an almost-universal sampler is indeed the best way to tackle the task of sampling. 

\subsection{Sampling from Single and Multi-model training}
We visualise the points sampled from single-task single-model (blue) and our proposed multi-task-multi-model training (red), as shown in Figure \ref{fig: vis2}. It is apparent that while there exists some overlapping points (which is intuitive as they are targeting the same task), the majority of points are different. Based on the accuracy improvements from our main paper, we can thus infer that our joint-training scheme allows us to capture features that are more universal towards a particular task.

\section{Reproducibility}
All our training (both meta-sampler and task adaptation) has a batch size of 24. Meta-training uses 5 gradient steps in the inner update. The rapid task adaptation uses the Adam optimiser with a learning rate of 1e-3. The ratio of task, simplification, and projection losses for task adaption is 1:1:1.

We provide the code for our meta-sampler trained on the three tasks, as well as the code for single-task multi-model training on SampleNet in our supplementary. Checkpoints and code will be made publicly available upon the publication of the work.